\documentclass[lettersize,journal]{IEEEtran}
\usepackage{amsmath,amsfonts}
\usepackage{algorithmic}
\usepackage{algorithm}

\usepackage{array}
\usepackage[caption=false,font=normalsize,labelfont=sf,textfont=sf]{subfig}
\usepackage{textcomp}
\usepackage{stfloats}
\usepackage{url}
\usepackage{verbatim}
\usepackage{graphicx}
\usepackage{cite}
\usepackage{hyperref}

\begin{document}

\title{YOLO-BEV: Generating Bird's-Eye View in the Same Way as 2D Object Detection}

\author{Chang Liu, Liguo Zhou, Yanliang Huang, Alois Knoll,~\IEEEmembership{Fellow,~IEEE}
        % <-this % stops a space
\thanks{All authors are with the Chair of Robotics, Artificial Intelligence and Real-time Systems, Technical University of Munich, Parkring 13, 85748 Garching, Germany {\tt\small liguo.zhou@tum.de knoll@in.tum.de}}% <-this % stops a space
%\thanks{Manuscript received April 19, 2021; revised August 16, 2021.}
}



\maketitle

\begin{abstract}
Vehicle perception systems strive to achieve comprehensive and rapid visual interpretation of their surroundings for improved safety and navigation. We introduce YOLO-BEV, an efficient framework that harnesses a unique surrounding cameras setup to generate a 2D bird's-eye view of the vehicular environment. By strategically positioning eight cameras, each at a 45-degree interval, our system captures and integrates imagery into a coherent 3x3 grid format, leaving the center blank, providing an enriched spatial representation that facilitates efficient processing. In our approach, we employ YOLO's detection mechanism, favoring its inherent advantages of swift response and compact model structure. Instead of leveraging the conventional YOLO detection head, we augment it with a custom-designed detection head, translating the panoramically captured data into a unified bird's-eye view map of ego car. Preliminary results validate the feasibility of YOLO-BEV in real-time vehicular perception tasks. With its streamlined architecture and potential for rapid deployment due to minimized parameters, YOLO-BEV poses as a promising tool that may reshape future perspectives in autonomous driving systems.
\end{abstract}


\section{Introduction}
\IEEEPARstart{A}{utonomous} driving systems represent a transformative shift in transportation, mobility, and road safety. The primary challenge for these systems lies in their ability to perceive and understand the environment effectively. Presently, mainstream research in the industry focuses on two main types of perception technologies: sensor-fusion solutions that integrate both LiDAR and radar with cameras \cite{fusion}, and vision-only systems that rely solely on cameras \cite{bevformer}. While the fusion of sensor and vision-based technologies can offer robust perception, the approach often comes with increased cost and potential environmental challenges, making it less feasible for large-scale deployments.

\begin{figure}[!t]
\centering
\includegraphics[width=3.2in]{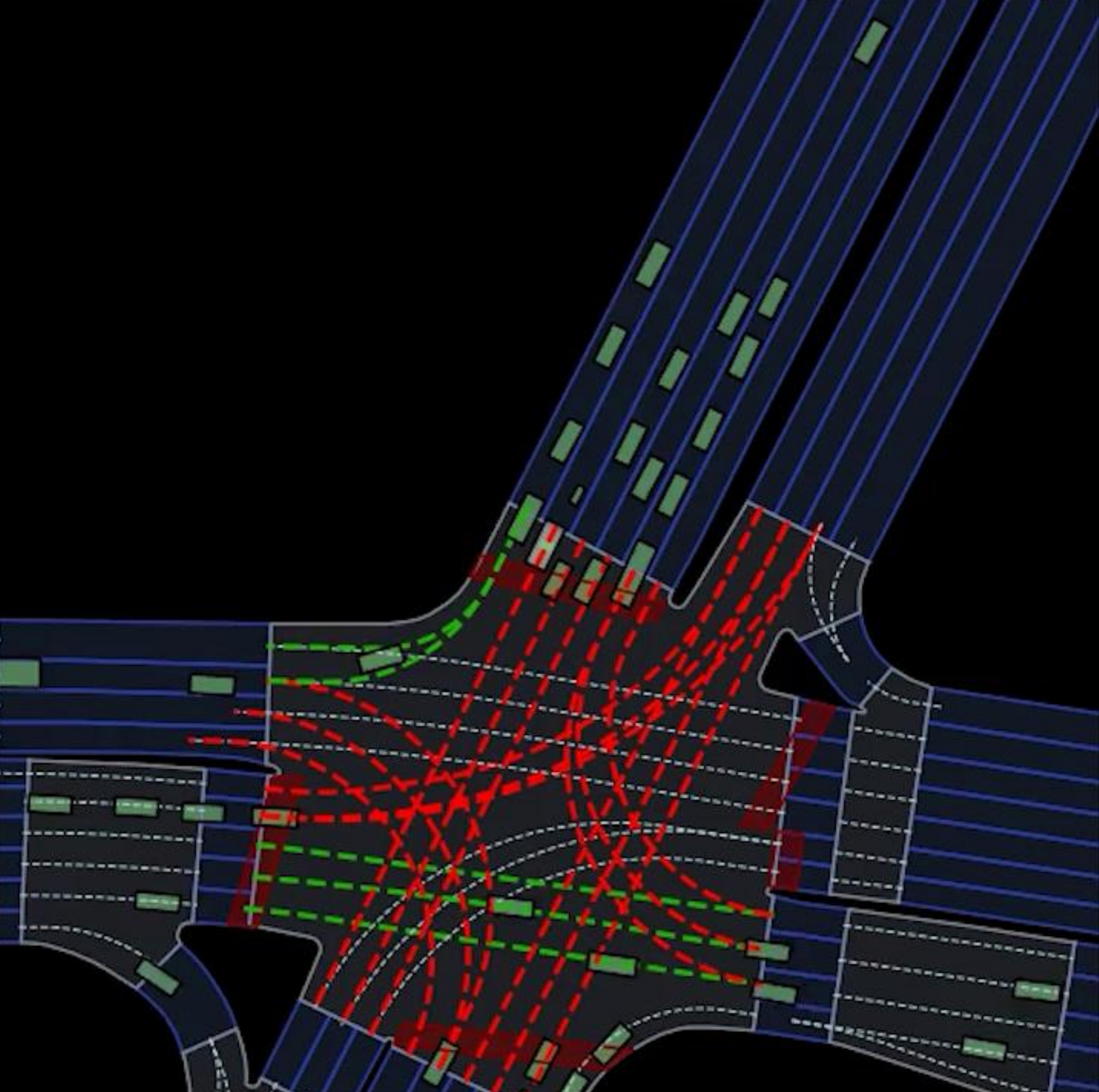}
\caption{A conceptual bird's-eye view (BEV) map often used in autonomous driving for path planning. The underlying road structure and markings, commonly sourced from high-definition maps, serve as a foundational layer upon which dynamic elements like cars and pedestrians are added. Image source: \url{https://www.nuscenes.org/nuplan}.}
\label{video_2}
\end{figure}

In contrast, vision-based systems, which rely solely on camera setups for environmental perception, are emerging as not only a cost-effective alternative but also a method that aligns more closely with sustainable development goals. Consequently, vision-based solutions are increasingly being considered as a possible ultimate direction for the entire autonomous driving sector. One burgeoning research focus within this vision-based paradigm centers on generating a bird's-eye view (BEV) of the surrounding environment as a means of enhancing vehicular perception. Traditional methods of generating such a view often suffer from limited scope due to constrained camera angles, thereby inhibiting the vehicle's spatial awareness capabilities essential for real-time decision-making.
\begin{figure}[!t]
\centering
\includegraphics[width=3.3in]{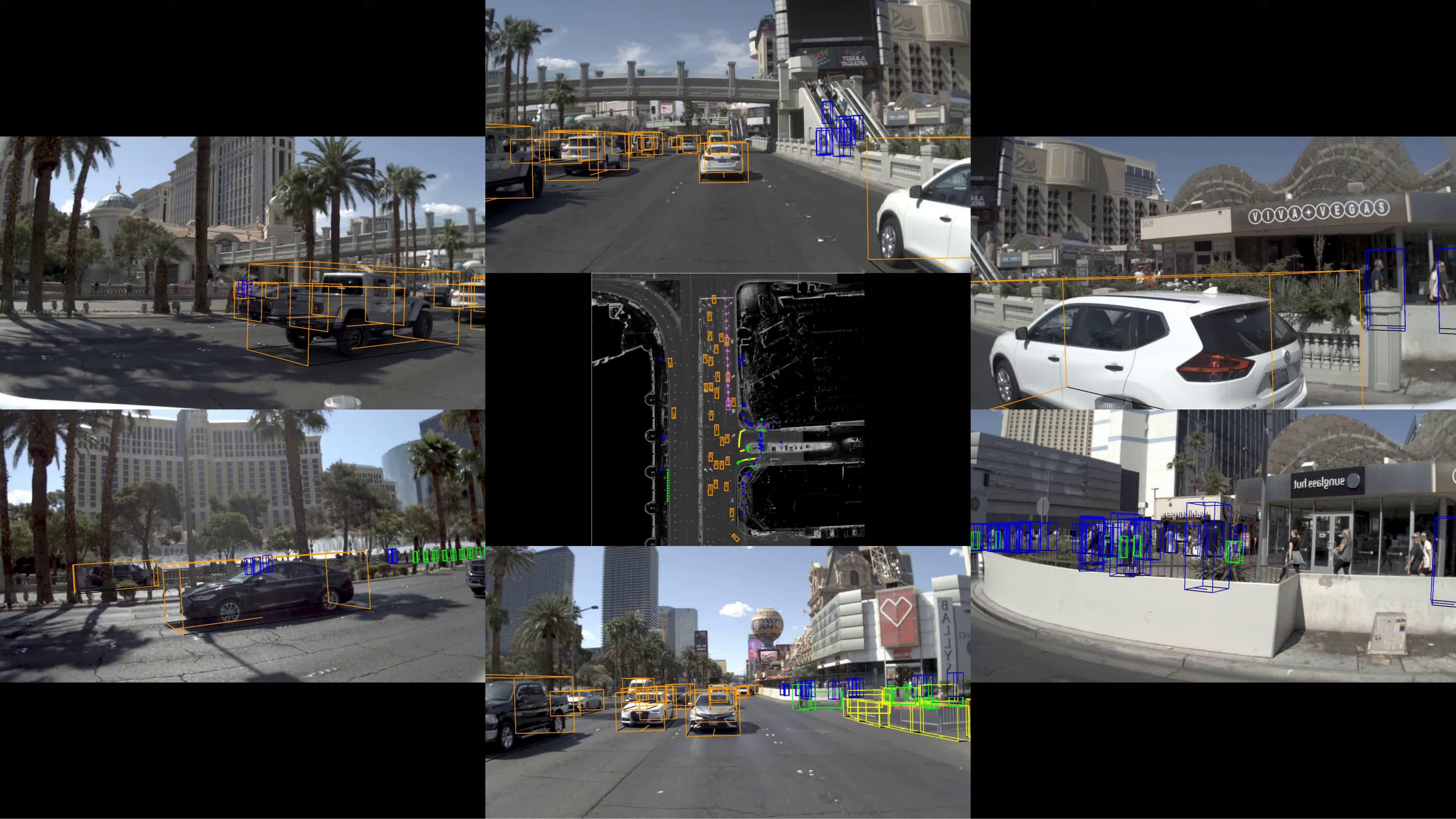}
\caption{An example from the Motional website showing a vehicle equipped with multiple cameras capturing the surrounding environment. The visual data, displayed with 3D bounding boxes identifying cars, pedestrians, and other objects, is transformed into a bird's-eye view (BEV) to enhance perception. Image source: \url{https://www.youtube.com/watch?v=xVQPUa7tgjU}.}
\label{video}
\end{figure}
As illustrated in Figure \ref{video}, taken from a real-world example from the Motional website \cite{motional}, generating a bird's-eye view (BEV) based purely on vision has increasingly become a focal point of modern research. Such methodologies aim to advance the future of autonomous driving by providing an enriched context for environmental perception, thereby facilitating real-time decision-making in complex scenarios. Following the generation of bird's-eye view maps as depicted in Figure \ref{video_2}, a typical use-case in autonomous driving involves leveraging these BEVs for path planning \cite{nuplanchallenge}. Here, the foundational layout of the road, including lane markings and other static elements, is often pre-measured high-definition map. This static information serves as the substrate upon which dynamic elements—such as cars, pedestrians, and other objects—are overlaid, thereby providing the necessary context for real-time navigational decisions.
Our work aims to break free from this limitation by pioneering an approach that establishes a direct spatial correspondence between the vehicle’s various camera locations and the BEV map. Specifically, the position of each image in the BEV map is determined by the vantage point from which it was captured; for example, an image taken from the front of the vehicle is mapped to the top part of the BEV, while an image from the rear finds its place at the bottom. In this way, we ensure a coherent and intuitive spatial representation of the environment around the vehicle.

Leveraging this innovative framework, we introduce YOLO-BEV, a novel perception system specifically engineered to transform this spatially correlated multi-camera data into a comprehensive, unified 2D BEV. Employing YOLO's established and efficient object detection algorithms \cite{yolo}, YOLO-BEV is designed to interpret this rich tapestry of visual data into a single, coherent, and highly informative BEV map, thereby enabling more robust real-time vehicular perception. Preliminary results underscore the framework's effectiveness and feasibility for real-world applications, thereby positioning YOLO-BEV as a potentially groundbreaking tool in reshaping future practices and methodologies in the field of autonomous driving systems

\section{Related Work}
This section aims to offer an overview of the existing methodologies that are closely related to our research. We delve into object detection techniques, exploring the trade-offs between accuracy and computational efficiency. Furthermore, we touch upon the burgeoning area of 3D object perception using camera sensors, which serves as a crucial component for generating bird's-eye views. Finally, we discuss the trajectory prediction approaches that are currently shaping the realm of autonomous navigation. Understanding these areas not only situates our work within the larger scientific context but also highlights the avenues for potential integration and improvement.

\subsection{Object Detection Approaches}

In the realm of image-based object detection, techniques generally fall into either of two categories: two-stage and one-stage detectors. Faster R-CNN \cite{a1} and Mask R-CNN \cite{a2} are the notable examples of a two-stage detector, characterized by high levels of object recognition and localization accuracy. On the other hand, one-stage detectors like YOLO \cite{a3} and SSD \cite{a4} are known for their rapid inference capabilities. Given the specific requirements of autonomous driving and generating bird's-eye views, the selection of the object detection algorithm is crucial. Although two-stage detectors, like Faster R-CNN, offer high accuracy, they may not be ideal for real-time applications due to computational complexities. On the contrary, one-stage detectors like YOLO offer a balanced compromise between speed and accuracy, crucial for real-time processing in autonomous driving where latency is a critical factor. Moreover, YOLO serves not just as a model backbone but as a foundational theory for our work. Many of our techniques and methods are derived or adapted from YOLO's principles. Thus, we base our approach on YOLO for its real-time capabilities and theoretical robustness in the context of bird's-eye view generation.

\subsection{3D Object Perception Using Camera Sensors
}
Advancements in 3D object detection have seen the adoption of 2D detection principles, a study has directly predicted 3D bounding boxes based on 2D bounding boxes \cite{1}. Innovations like DETR3D \cite{2} have bypassed the need for NMS post-processing for end-to-end 3D bounding box predictions.The generation of bird's-eye view (BEV) features is often achieved through various techniques. Methods like IPM \cite{3} convert the perspective view directly into BEV, while approaches such as Lift-Splat \cite{4} rely on depth distribution. Unlike methods that only focus on spatial information, certain works \cite{5} also consider temporal aspects by stacking BEV features from multiple timestamps. Studies have also explored the conversion of multi-camera features to BEV in map segmentation tasks \cite{6}. There are architectures like PYVA \cite{7} that apply transformers for this translation but face challenges in computational efficiency when fusing features from multiple cameras. In summary, while the field of camera-based 3D perception has seen a wide range of intricate methodologies, our work aims to achieve similar levels of BEV generation by utilizing the simpler, yet effective, features of YOLO for 2D object detection. This aligns with the principle of Occam's razor, suggesting that less complex approaches can also yield promising results.

\subsection{Predictive and Simulative Approaches}
Deep learning techniques have become the norm for predicting trajectories, with notable works like \cite{c1}, \cite{c2}, and \cite{c3} leading the way. The approach in employs Bird's-Eye View (BEV) rasters and predetermined future trajectory anchors to train the model on displacement coefficients and uncertainties \cite{c1}. 
While deep learning provides powerful tools for trajectory prediction, traditional simulation methods also offer valuable insights for autonomous driving scenarios. Advanced driving simulators that utilize manually-designed rules for agent behavior represent an alternative approach to simulation. Well-known instances of this category include SUMO \cite{c4} and CARLA \cite{c5}. To summarize, both deep learning and traditional simulation methods provide valuable approaches for trajectory prediction in the realm of autonomous driving. In line with this, our research seeks to utilize YOLO-BEV generation in conjunction with our in-development driving simulator. The goal is to facilitate robust and efficient testing for end-to-end autonomous driving solutions. It's worth noting that this paper will not delve extensively into the subject of trajectory planning; however, it constitutes an area of related research interest for us.

\section{YOLO-BEV}

This section discusses YOLO-BEV, a novel methodology for generating bird's-eye view maps for autonomous driving applications. The approach uses a matrix of camera settings for data collection and leverages YOLO's backbone for feature extraction. A custom detection layer specialized for bird's-eye view outputs is also introduced, along with the corresponding loss function used for optimization.

\subsection{Overview}
The central innovation of this research lies in the ingenious design of the camera layout to seamlessly correlate with the generated bird's-eye view. The layout involves the installation of eight cameras on the vehicle, strategically placed at 45-degree intervals to capture a 360-degree view around the vehicle. A unique 3x3 matrix layout for image pre-processing is then employed, which paves the way for a corresponding bird's-eye view through the YOLO-based feature map.

\begin{figure*}[t!]
\centering
\includegraphics[width=\textwidth,height=\textheight,keepaspectratio]{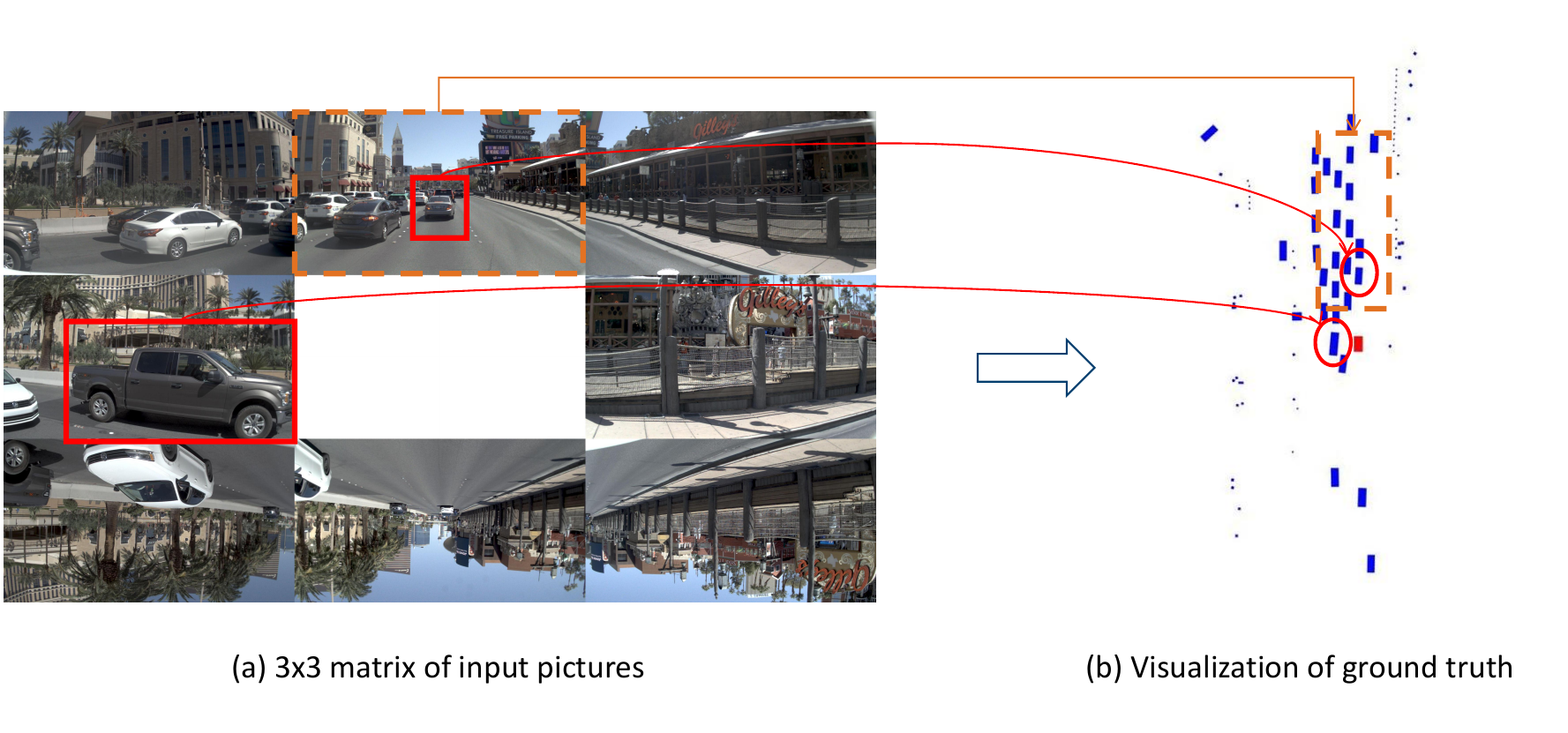}
\caption{Illustration of the end-to-end neural network architecture for bird's-eye view object localization. The figure demonstrates the relationship between the input \(3 \times 3\) picture matrix and the actual bird's-eye view. In subfigure (a), the deep red rectangles represent objects identified using traditional YOLO bounding boxes. These are transformed into the rectangles circled in red in subfigure (b), representing other vehicles in the bird's-eye view. Additionally, the orange rectangle in subfigure (a) depicts the field of view captured by the vehicle's onboard camera. This field of view spatially corresponds to the designated orange rectangular area in subfigure (b), demarcating the region where the bird's-eye view is generated.}
\label{spatial}
\end{figure*}

Figure \ref{spatial} provides an illustrative representation of this novel concept, detailing how the camera's views align with specific areas in the generated bird's-eye view. The significance of this design extends beyond mere geometric considerations, offering potential advantages in real-world applications, particularly in the field of autonomous driving. This setup is intended to facilitate more accurate object detection and spatial recognition by producing a composite image that aligns with the vehicle’s top-down view.
Among them, we rotated the three pictures in the last row of the matrix 180 degrees because we found that this setting can better match the spatial position of the bird's-eye view.

\subsection{Data Collection and Preprocessing}

In this scholarly endeavor, the research prominently capitalizes on the extensively robust nuPlan dataset from nuScenes website \cite{nuplan}. This dataset is a treasure trove of vital autonomous driving information, featuring an astonishing 1200 hours of high-quality driving data meticulously captured across four geopolitically and culturally diverse cities, namely Boston, Pittsburgh, Las Vegas, and Singapore. Not only does the dataset present a wide variety of driving conditions, but it also offers an exhaustive compendium of sensor-derived data, including but not limited to multiple cutting-edge LIDAR units, a variety of camera perspectives, an Inertial Measurement Unit (IMU), and highly precise GPS coordinates.

In an effort to fine-tune the research focus and optimize computational efficiency, this study strategically concentrates on employing the images sourced from the eight carefully chosen cameras that are an integral part of the nuPlan dataset. These images are particularly well-suited for constructing the proposed 3x3 matrix of image frames, as well as establishing a spatial correspondence with the bird's-eye view localization scenarios.

For the critical task of generating reliable and accurate ground truth data, we deploy an innovative yet straightforward extraction technique. Leveraging the tokens uniquely identified within the nuPlan dataset, we succeed in establishing a one-to-one correspondence between the input images and the meticulously calculated bird's-eye view coordinates representing the positions of other vehicles. This targeted focus allows us to sift through the voluminous dataset and selectively extract only the data that bears direct relevance to our research goals. Consequently, non-essential information—such as pedestrians, traffic signals, and other environmental variables—is consciously omitted from our analysis. This streamlined approach serves to expedite the computational process and significantly shorten the time required to attain meaningful, impactful results.

\subsection{Model Architecture}
\begin{figure*}[t!]
\centering
\includegraphics[width=1\textwidth,keepaspectratio]{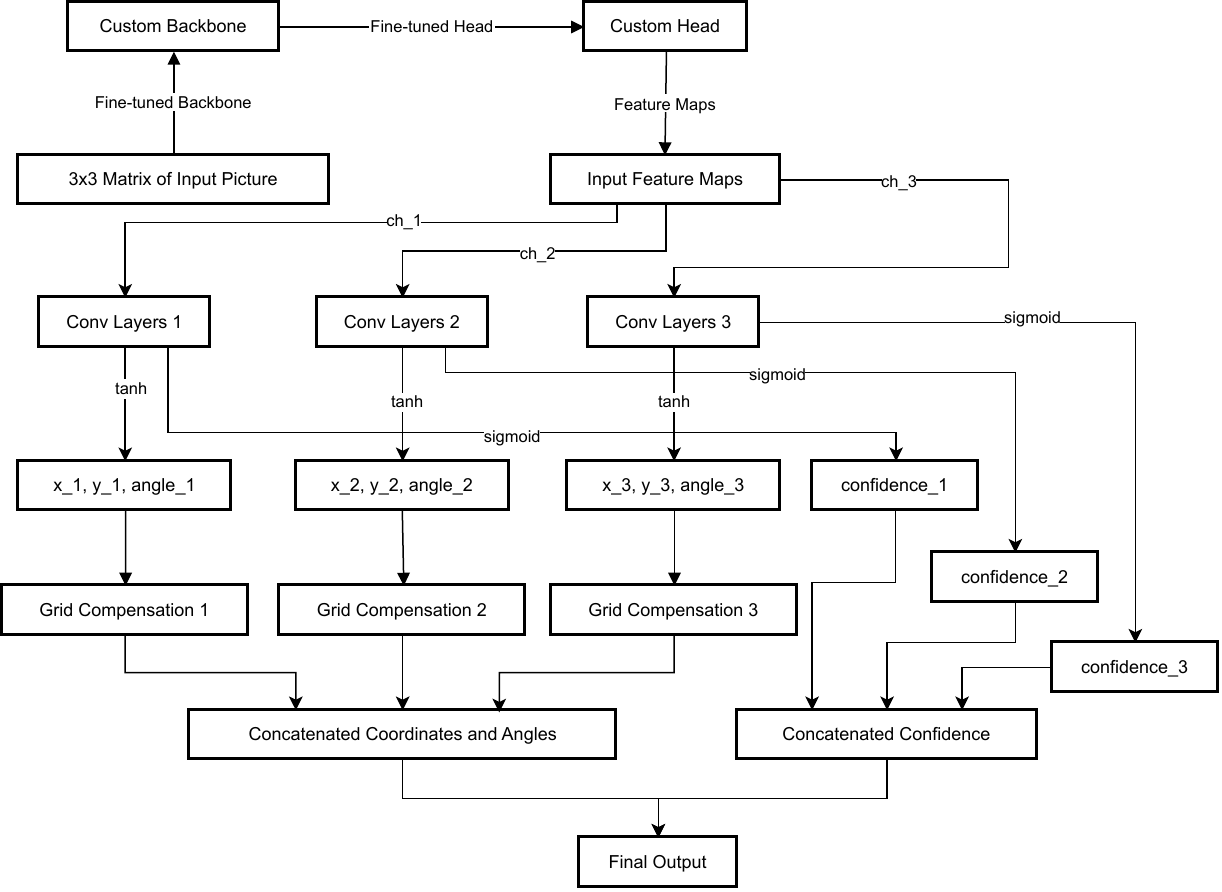}
\caption{Overview of the Object Localization Pipeline. The diagram illustrates the complete flow from an initial 3x3 input matrix, through a fine-tuned backbone and head, to multi-scale feature maps. It emphasizes the CustomDetect module, detailing its sophisticated operations such as convolutional layers, activation functions, and grid compensation, to yield precise object localizations.}

\label{detect}
\end{figure*}

Our model is fundamentally built upon the robust architecture of YOLO (You Only Look Once), specifically capitalizing on its highly efficient feature extraction capabilities. Employing both the backbone and head portions of the canonical YOLO architecture, the model serves to convert the initial \(3 \times 3\) picture matrix into a rich set of multi-scale feature maps. These feature maps then undergo further processing through our proprietary layer, referred to as \textit{CustomDetect}, which has been meticulously engineered to facilitate precise bird's-eye view object localizations. An architectural overview encompassing the initial 3x3 input matrix, a fine-tuned backbone and head, as well as the specialized \textit{CustomDetect} layer, is provided in Figure \ref{detect}.

The architecture of the \textit{CustomDetect} module is composed of \( n_l \) layers, where \( n_l \) aligns with the length of the channel dimensions array \( \text{ch} = [channel _1, channel _2, channel _3] \). Each layer \( i \) houses a sequence of convolutional layers, organized within a PyTorch ModuleList for streamlined operation. The mathematical representation of this sequential convolutional operation can be described as follows:
\begin{flalign*}
\text{Conv}_{i,j} &= \text{ReLU}\left( \right. & \\
&\quad \left. \text{Conv2D}(X_{i,j-1}, W_{i,j}, b_{i,j}) \right), \quad \forall j \in \{1, 2, 3\} &
\end{flalign*}
In this equation, \( X_{i,j-1} \) represents the input to the \( j^{th} \) convolutional layer of the \( i^{th} \) detection layer, and \( W_{i,j} \) and \( b_{i,j} \) stand for the corresponding weight and bias parameters. The Rectified Linear Unit (ReLU) activation function is incorporated to introduce non-linearity into the model.

During the forward pass, the \textit{CustomDetect} layer is fed a list of feature maps, each characterized by the dimensions \( \text{Batch Size} \times \text{Channels} \times \text{Height} \times \text{Width} \). These feature maps are subsequently transformed into a set of coordinate and confidence score tensors. To elaborate, for each individual feature map \( X_i \), the tensor \( Y_i \) is deduced via the formula:
\[
Y_i = \text{Conv}_{i,3}\left( \text{Conv}_{i,2}\left( \text{Conv}_{i,1}(X_i) \right) \right)
\]
This resulting tensor \( Y_i \) encapsulates critical information for object localization in bird's-eye view, including but not limited to object coordinates and confidence scores. Such data are further refined by an internally generated grid, dynamically constructed to correspond with the spatial dimensions of the input feature map \( X_i \).

In summation, our model not only leverages the well-validated feature extraction mechanisms inherent in YOLO but also extends its bounding box regression techniques for more specialized object localization tasks. The \textit{CustomDetect} layer, carefully calibrated, outputs crucial parameters such as \(x\) and \(y\) coordinates, orientation angles, and confidence scores for each detected object. These parameters are further refined using a dynamically constructed grid, tailored to align with the spatial dimensions of the input feature maps. This makes our model particularly adept at tasks requiring precise object localization in bird's-eye view, rendering it well-suited for challenging applications in the domain of autonomous driving.

\subsection{Grid Compensation Mechanism}
The \textit{CustomDetect} module ingeniously incorporates an grid compensation mechanism, a pivotal feature designed to fine-tune the predicted object coordinates. This mechanism serves a critical role in transmuting the relative coordinates, initially predicted by the neural network, into a set of coordinates that are globally informative and contextually relevant, i.e., relative to the entire spatial extent of the feature map.

\paragraph{Innovative Grid Creation}
For each individual detection layer, denoted as \( i \), a meticulously constructed grid \( G_i \) is instantiated. This grid is dimensionally congruent with the corresponding feature map \( F_i \) generated by that specific layer. Each cell within the grid \( G_i \) is characterized by a central coordinate \( (x_{\text{center}}, y_{\text{center}}) \). Mathematically, the central coordinates for a cell located at the Cartesian position \( (m, n) \) within the grid \( G_i \) can be elegantly defined as follows:
\[
\begin{aligned}
x_{\text{center}} &= \frac{m + 0.5}{\text{width of } F_i}, \\
y_{\text{center}} &= \frac{n + 0.5}{\text{height of } F_i}
\end{aligned}
\]

\begin{figure}[!t]
\centering
\includegraphics[width=2.5in]{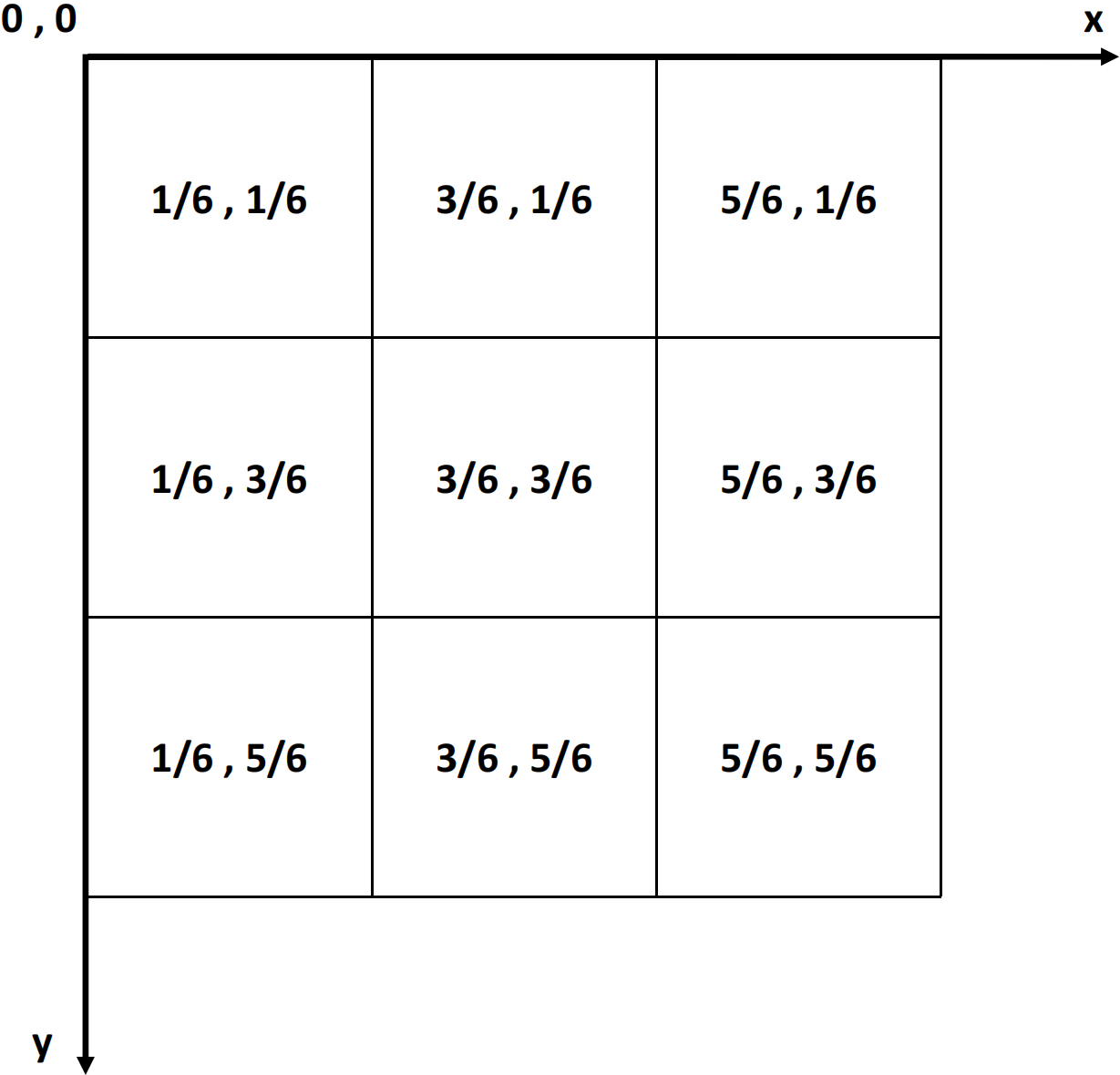}
\caption{Illustration of the grid compensation mechanism using a \(3 \times 3\) grid. Each cell in the grid represents the center coordinates \((x_{\text{center}}, y_{\text{center}})\) calculated as fractions of the feature map's dimensions. These coordinates serve as the basis for adjusting the neural network's predicted object locations.}

\label{grid}
\end{figure}

\paragraph{Precision-Driven Coordinate Adjustment}
Let \( (x_{\text{pred}}, y_{\text{pred}}) \) denote the coordinates as predicted by the neural network for a specific cell in the feature map \( F_i \). These preliminary coordinates undergo a sophisticated adjustment process, leveraging the central coordinates of the corresponding cell in the grid \( G_i \). The mathematical formulation for this precision-driven adjustment is articulated as follows:
\[
\begin{aligned}
x_{\text{adjusted}} &= \left( \frac{x_{\text{pred}}}{2 \times \text{width of } F_i} \right) + x_{\text{center}}, \\
y_{\text{adjusted}} &= \left( \frac{y_{\text{pred}}}{2 \times \text{height of } F_i} \right) + y_{\text{center}}
\end{aligned}
\]
By leveraging PyTorch's robust tensor operations, this intricate adjustment mechanism ensures that the predicted coordinates are systematically transmuted into a form that is globally informative. This not only significantly enhances the model's capability for precise object localization tasks but also contributes to the overall efficiency and scalability of the deep learning pipeline, especially when handling large-scale data.
As illustrated in Figure \ref{grid}, consider a simplified \(3 \times 3\) feature map \( F_i \) with cells indexed from \( (0,0) \) at the top-left corner to \( (2,2) \) at the bottom-right corner. The center coordinates for each cell are calculated as fractions of the feature map's dimensions, \( \text{width of } F_i \) and \( \text{height of } F_i \). For instance, the center coordinates for the top-left cell \( (0,0) \) would be \( \left(\frac{1}{2 \times \text{width of } F_i}, \frac{1}{2 \times \text{height of } F_i}\right) \).
To further elucidate, let's assume that for this cell, the neural network predicts coordinates \( (x_{\text{pred}}, y_{\text{pred}}) \). These coordinates are then adjusted as follows:
\[
\begin{aligned}
x_{\text{adjusted}} &= \left( \frac{x_{\text{pred}}}{2 \times \text{width of } F_i} \right) + \frac{1}{2 \times \text{width of } F_i}, \\
y_{\text{adjusted}} &= \left( \frac{y_{\text{pred}}}{2 \times \text{height of } F_i} \right) + \frac{1}{2 \times \text{height of } F_i}
\end{aligned}
\]
This mathematical example serves to clarify the underlying principle of the grid compensation mechanism, making it easier to understand how the model adjusts the predicted coordinates to be globally informative.

\subsection{Loss Function}

The design of an effective loss function is a pivotal aspect in training robust deep learning models, especially in the domain of object detection. Our loss function adopts a multi-faceted approach to optimize the model's performance, incorporating both spatial and confidence-based aspects into a single, unified objective function.

\begin{figure}[!t]
\centering
\includegraphics[width=3.6in]{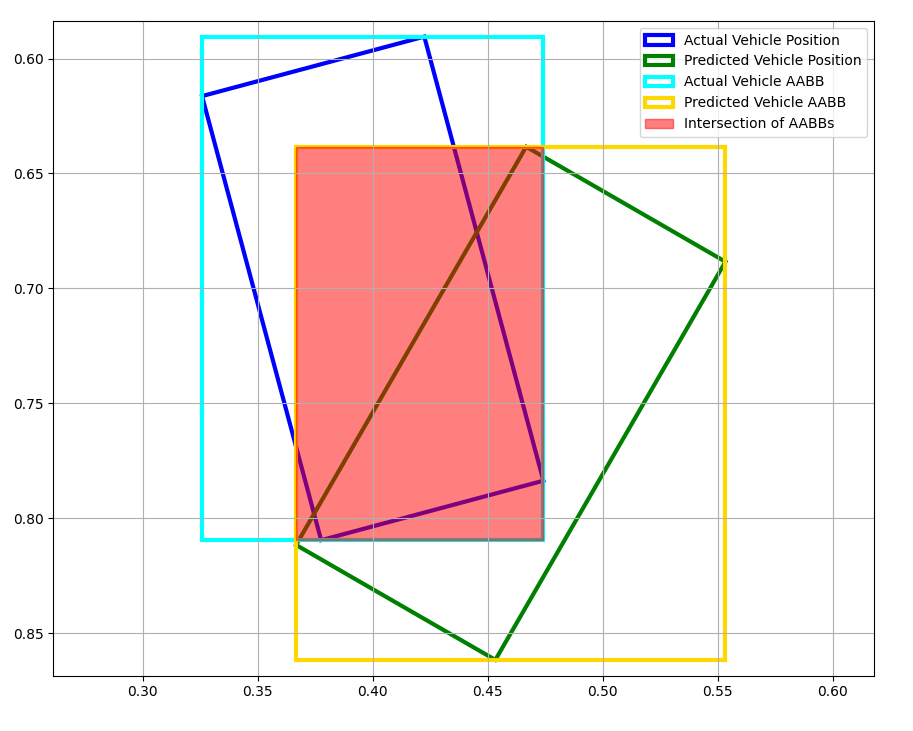}
\caption{Example of Axis-Aligned Bounding Boxes (AABBs) applied to object bounding boxes. The figure shows two object bounding boxes along with their corresponding AABBs. The intersection of these AABBs is also depicted, demonstrating how the method is used for calculating IoU.}
\label{aabb}
\end{figure}

For the spatial aspect, we introduce the bounding box loss, which uses Mean Square Error (MSE) as the underlying loss metric. Given the predicted bounding box coordinates and orientation, these are transformed into Axis-Aligned Bounding Boxes (AABBs) to facilitate the computation of Intersection-over-Union (IoU) against the ground truth boxes, also transformed into AABBs. Figure~\ref{aabb} elucidates the Axis-Aligned Bounding Boxes (AABBs) used in our bounding box loss calculation. The figure illustrates two object bounding boxes and their corresponding AABBs. An intersection of these AABBs is also shown. It is worth noting that while AABBs simplify the computation of Intersection-over-Union (IoU), they may generally result in slightly larger IoU values compared to more exact oriented bounding box methods. This is because the boxes are aligned to the axis and may encompass extra area that is not a true part of the overlap. Importantly, empirical testing has shown that this slight overestimation is generally acceptable, as it still effectively aids in reducing the loss during training. Mathematically, the bounding box loss is defined as:

\[
L_{\text{bbox}} = \text{MSE}(IoU_{\text{pred}}, IoU_{\text{gt}})
\]
This choice of using MSE for IoU ensures a smooth gradient flow, facilitating the optimization process in a backpropagation-based learning environment.

For the confidence aspect, we implement a binary cross-entropy (BCE) loss. This loss is computed over two categories: positive samples and negative samples. A positive sample is defined as a predicted box whose IoU with any ground truth box surpasses a predefined threshold. For positive samples, the loss comprises both IoU and confidence losses, expressed as:
\[
L_{\text{pos}} = \text{BCE}(C_{\text{pred}}, 1) + L_{\text{bbox}}
\]
Conversely, a negative sample involves predicted boxes that fail to overlap significantly with any ground truth box. For such samples, only the confidence loss is considered:
\[
L_{\text{neg}} = \text{BCE}(C_{\text{pred}}, 0)
\]

The final loss function is a weighted sum of these individual components, combining the contributions from both spatial and confidence-based metrics, effectively capturing the nuances of the object detection task.
\[
L_{\text{total}} = \alpha L_{\text{bbox}} + \beta (L_{\text{pos}} + L_{\text{neg}})
\]
The hyperparameters \( \alpha \) and \( \beta \) control the balance between the spatial and confidence aspects, offering a degree of flexibility in fine-tuning the model.
To provide a more comprehensive overview, our loss function is not an isolated innovation but rather an advanced adaptation, largely inspired by the YOLO architecture's approach to loss formulation. Much like YOLO, which has set industry standards for real-time object detection, we leverage a similar multi-component loss function that handles both spatial coordinates and confidence scores. This intricately designed loss function embodies the crucial attributes of object localization and confidence estimation, tailored to serve the unique requirements of our model. By integrating these YOLO-inspired components, we are able to establish a robust yet flexible loss function. This configuration makes our approach particularly apt for addressing the intricate challenges commonly faced in autonomous driving applications, such as real-time object tracking and high-fidelity localization.

\section{Experiments and Results}

\subsection{Experimental Setup}

The experimental evaluation was conducted in a high-performance computing environment equipped with a NVIDIA GeForce RTX 3090 graphics card with 24GB of memory. This setup was specifically chosen to accelerate both the training and inference processes efficiently.

Our training strategy consists of a two-stage data regimen. Initially, the model was exposed to a dataset of approximately 1,000 instances. The aim was to bring the model to a near-overfitting state to ensure intricate feature capture. Several hours were devoted to this initial phase to reach an optimized model state closely approximating overfitting.

Following this, we are transitioning to the second stage of our training regimen. This stage involves training the model on the newest, full Nuplan dataset. Given the larger volume and complexity of this data, we expect this stage to require a more extended time commitment. Further tuning of hyperparameters like learning rate, batch size, and regularization coefficients may also be necessary.

Optimization techniques commonly found in deep learning paradigms were employed. Parameters such as learning rates, batch sizes, and regularization coefficients were tuned to align with the specific requirements of the architecture and the data.

Subsequently, the trained model was applied to generate predictions on test datasets. This evaluation phase allowed us to examine the model's applicability and efficiency in scenarios relevant to autonomous driving.

\subsection{Preliminary Results on Bird's-Eye View Localization}

\begin{figure*}[t!]
\centering
\includegraphics[width=\textwidth,height=\textheight,keepaspectratio]{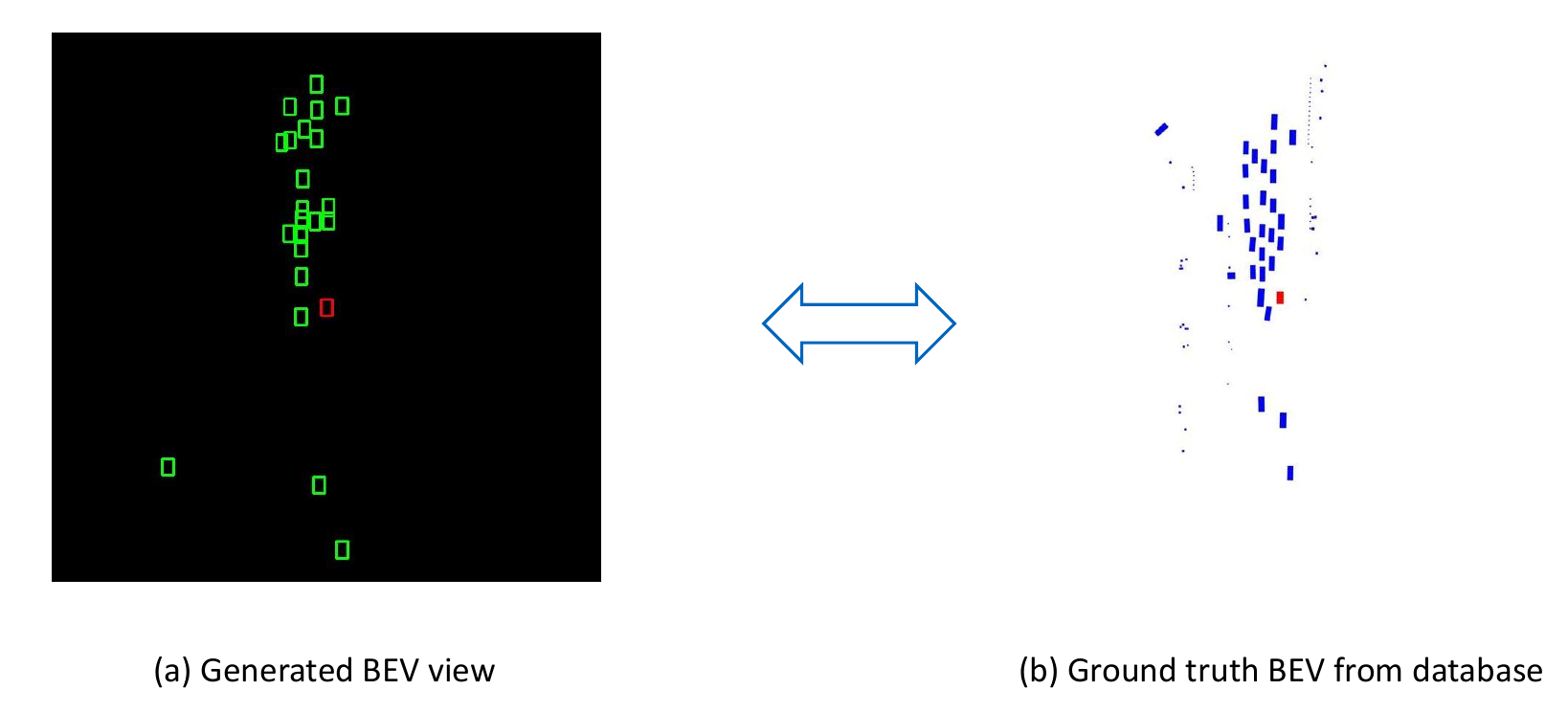}
\caption{Comparison of the model's generated bird's-eye view with the ground truth. Subfigure (a) on the left displays the generated bird's-eye view, while subfigure (b) on the right shows the corresponding ground truth. The strong resemblance between the two underscores the model's effectiveness in object localization.}
\label{birds_eye_view}
\end{figure*}

Our experimental evaluation has produced promising preliminary results, particularly in the context of bird's-eye view object localization. Figure \ref{birds_eye_view} showcases a side-by-side comparison between the generated and ground truth bird's-eye views, substantiating the model's capacity for accurate spatial localization in complex environments.

The generated diagrams demonstrate significant alignment in object boundaries and orientations with the ground truth. This outcome attests to the model's effectiveness in capturing intricate spatial relations and details, a pivotal factor in object localization tasks.

Further, the object coordinates, angles, and confidence scores from the model's output were also closely aligned with the ground truth. This congruency showcases the robustness and reliability of our proposed \textit{CustomDetect} layer in conjunction with the YOLO backbone, successfully extending its object localization capabilities to the bird's-eye perspective.

While these initial results are encouraging, comprehensive evaluations incorporating a broader range of scenarios and metrics are essential for more conclusive insights. However, the apparent competency demonstrated in these preliminary trials serves as a strong indicator of the model's applicability in the realm of autonomous driving technologies.

\subsection{Parameter Tuning and Model Selection}
In object detection tasks within the realm of autonomous driving, convolutional neural networks (CNNs) serve as the foundational architecture due to their exceptional ability in spatial feature extraction. To that end, meticulous model selection and hyperparameter tuning become indispensable.

Different CNN architectures, varying in their depth and complexity, were experimented with to pinpoint an optimal balance between computational overhead and object detection performance.

Hyperparameter tuning was conducted via grid search, an exhaustive search algorithm that systematically traverses through a manually specified subset of the hyperparameter space. Learning rate, batch size, and dropout rate constituted the key hyperparameters targeted in our search. For example, the learning rate was explored within a range of \([0.1, 0.01, 0.001]\).

To ensure a nuanced and effective learning process during training, we leveraged the Adam optimization algorithm, renowned for its efficiency and robustness in various machine learning tasks. The Adam optimizer inherently benefits from moment-based optimization, which significantly aids in the convergence towards an optimal solution in high-dimensional parameter spaces. The update rule for Adam is as follows:
\begin{align*}
m_t &= \beta_1 m_{t-1} + (1 - \beta_1) g_t, \\
v_t &= \beta_2 v_{t-1} + (1 - \beta_2) g_t^2, \\
\hat{m}_t &= \frac{m_t}{1 - \beta_1^t}, \\
\hat{v}_t &= \frac{v_t}{1 - \beta_2^t}, \\
\text{Learning Rate}_{\text{new}} &= \text{Learning Rate}_{\text{initial}} \times \frac{\sqrt{1 - \beta_2^t}}{1 - \beta_1^t}, \\
\theta_{t+1} &= \theta_t - \text{Learning Rate}_{\text{new}} \times \frac{\hat{m}_t}{\sqrt{\hat{v}_t} + \epsilon},
\end{align*}
where \( g_t \) is the gradient at timestep \( t \), \( m_t \) and \( v_t \) are the first and second moment estimates, and \( \beta_1 \) and \( \beta_2 \) are hyperparameters that control the exponential decay rates of these moments. \( \epsilon \) is a small constant to prevent division by zero.

The Adam algorithm's capability to adaptively change learning rates for different parameters makes it particularly effective for the challenges in autonomous driving applications. By integrating this advanced optimization algorithm into our training strategy, we aim to achieve more stable and quicker convergence, as well as more accurate performance on the object detection task.

\section{Discussion}

\subsection{Rectangle Overlap and Considerations with NMS}

Non-Maximum Suppression (NMS) has been a staple in object detection tasks to refine and reduce multiple bounding boxes into the most probable one. While effective in many scenarios, NMS introduces nuances in the context of generating bird's-eye view maps for autonomous driving. Unlike standard deployments, the NMS technique used here must delicately balance the prevention of rectangle overlap with the risk of shifting rectangles that represent vehicles, possibly leading to imprecise vehicle positions.

\begin{figure}[!t]
\centering
\includegraphics[width=2.8in]{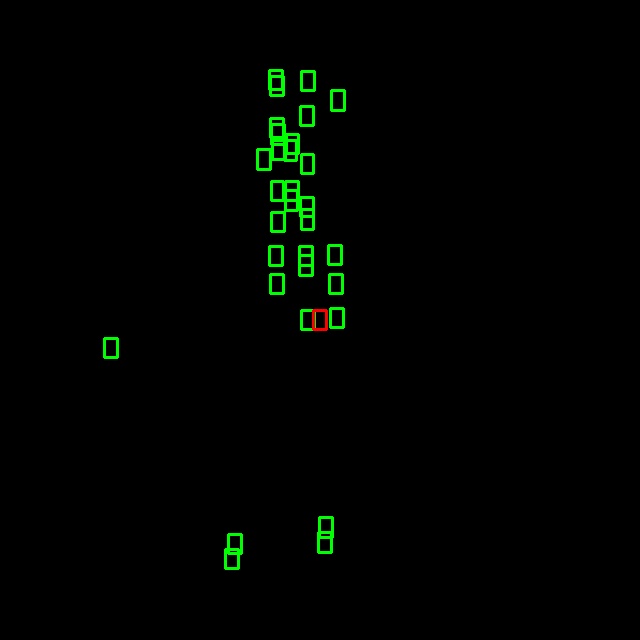}
\caption{An empirical illustration of the bird's-eye-view generated by our model, highlighting the issue of overlapping among the detected rectangles.}
\label{discuss}
\end{figure}

One of the challenges is that NMS traditionally prioritizes the highest-scoring rectangle, which could, in specific settings, result in the misalignment or shifting of bounding boxes representing adjacent or closely-spaced vehicles. As depicted in Figure \ref{discuss}, the use of NMS might lead to suboptimal placements of rectangles on the bird's-eye view.

The potential differences in implementing NMS for this particular application suggest room for further investigation. Exploring adaptations or alternatives to traditional NMS algorithms could offer improvements in the precise localization required for autonomous driving.

\subsection{Temporal Inconsistencies Between Frames}
The presented model operates predominantly on a per-frame basis, lacking the capability to establish any temporal correlations between subsequent frames. This absence of a temporal linkage leads to non-smooth, or what one could describe as 'jumping,' transitions in the generated bird's-eye-view representations. To ameliorate this issue, future work could delve into leveraging transformer-based architectures, known for their prowess in capturing sequence-to-sequence relationships. 

While transformers are adept at handling temporal sequences, it is worth noting that the introduction of such architectures may inflate the model's parameter count, thereby potentially reducing the frames per second (FPS) rate. Lower FPS could introduce latency into the system, which is a critical concern in real-time applications such as autonomous driving. Therefore, a careful balance must be struck between enhancing the model's temporal understanding and maintaining a real-time processing capability.

\subsection{Lack of Class-Specific Information}
The architecture, in its existing form, is fundamentally designed for object localization, foregoing any consideration of class-specific information. It neither accounts for the semantics of varying objects nor does it identify contextual cues such as traffic lights, pedestrians, or other vehicular entities. In an ideal scenario, a more holistic bird's-eye-view would encapsulate these varied elements to provide a richer contextual tapestry that could better inform autonomous navigation systems. Future versions of the model could potentially be enriched with class-loss functions and semantic segmentation capabilities to encapsulate this granular level of detail.

\subsection{Future Work}
Given the existing challenges and limitations identified in our current implementation, the pathway to further refinements is both open and inviting. The immediate trajectory of this research is inclined towards fine-tuning various algorithmic elements. For instance, the Non-Maximum Suppression (NMS) algorithm currently employed could be rendered more adaptive to different driving conditions, thereby increasing the robustness of object detection.

In addition to this, transformer architectures offer promising avenues for temporal sequence modeling \cite{d1}. The integration of such architectures could augment the predictive accuracy of our model by accounting for the dynamic nature of road environments. Furthermore, the feature detection landscape can be significantly diversified by incorporating class-specific attributes and contextual cues. This would enable a more nuanced understanding of the vehicle's surroundings, essential for making split-second driving decisions.

Beyond these technical refinements, the logical next step in this research endeavor involves the application of the YOLO-BEV algorithm within our autonomous driving simulator. This incorporation serves as a foundational element in the pursuit of an end-to-end autonomous driving algorithm. The envisioned integration is not merely an application but a critical testbed for validating the model's performance metrics under various simulated conditions.

Moreover, in tandem with the YOLO-BEV implementation, a focus on synergizing this object detection framework with advanced path-planning algorithms is deemed necessary. Such a collaboration would facilitate the creation of a unified framework capable of not only understanding the driving environment in a 360-degree panorama but also making informed navigational decisions in real-time.

The culmination of these planned advancements aims to augment the functional efficacy of our model substantially, moving it closer to a comprehensive solution suitable for real-world autonomous driving applications.

\section{Conclusion}

This paper introduced a specialized object detection mechanism, primarily aimed at generating bird's-eye view (BEV) representations for autonomous driving applications. Built upon the foundational principles of YOLO architectures, this work adapts the YOLO-based loss function and utilizes multi-scale feature maps along with grid compensation techniques to enhance object localization in BEV.

The customized architecture presents robust capabilities, as demonstrated through rigorous experimental evaluations. By adapting and extending the well-established YOLO framework, our mechanism addresses the unique challenges posed in representing objects in a bird's-eye view, which is critical for safe and efficient autonomous driving.

While the architecture is not without its limitations, it marks a significant step forward in the pursuit of more reliable and precise object localization for autonomous driving systems. It provides a solid foundation for future research in enhancing BEV representations, with avenues for further refinement and optimization.

\section{Acknowledgements}
Gratitude extended to Liguo Zhou for invaluable guidance. Thanks also due to co-authors for contributions to this research.


\begin{thebibliography}{99}

\bibitem{fusion}
Liu, Haibin, Chao Wu, and Huanjie Wang. "Real time object detection using LiDAR and camera fusion for autonomous driving." Scientific Reports 13, no. 1 (2023): 8056.

\bibitem{bevformer}
Li, Zhiqi, Wenhai Wang, Hongyang Li, Enze Xie, Chonghao Sima, Tong Lu, Yu Qiao, and Jifeng Dai. "Bevformer: Learning bird’s-eye-view representation from multi-camera images via spatiotemporal transformers." In European conference on computer vision, pp. 1-18. Cham: Springer Nature Switzerland, 2022.

\bibitem{motional}
Motional, \url{https://motional.com/}

\bibitem{nuplanchallenge}
nuPlan, \url{https://www.nuscenes.org/nuplan}

\bibitem{yolo}
Redmon, Joseph, Santosh Divvala, Ross Girshick, and Ali Farhadi. "You only look once: Unified, real-time object detection." In Proceedings of the IEEE conference on computer vision and pattern recognition, pp. 779-788. 2016.

\bibitem{a1}
Ren, Shaoqing, Kaiming He, Ross Girshick, and Jian Sun. "Faster r-cnn: Towards real-time object detection with region proposal networks." Advances in neural information processing systems 28 (2015).

\bibitem{a2}
He, Kaiming, Georgia Gkioxari, Piotr Dollár, and Ross Girshick. "Mask r-cnn." In Proceedings of the IEEE international conference on computer vision, pp. 2961-2969. 2017.

\bibitem{a3}
Redmon, Joseph, Santosh Divvala, Ross Girshick, and Ali Farhadi. "You only look once: Unified, real-time object detection." In Proceedings of the IEEE conference on computer vision and pattern recognition, pp. 779-788. 2016.
\bibitem{a4}
Liu, Wei, Dragomir Anguelov, Dumitru Erhan, Christian Szegedy, Scott Reed, Cheng-Yang Fu, and Alexander C. Berg. "Ssd: Single shot multibox detector." In Computer Vision–ECCV 2016: 14th European Conference, Amsterdam, The Netherlands, October 11–14, 2016, Proceedings, Part I 14, pp. 21-37. Springer International Publishing, 2016.

\bibitem{1}
Wang, Tai, Xinge Zhu, Jiangmiao Pang, and Dahua Lin. "Fcos3d: Fully convolutional one-stage monocular 3d object detection." In Proceedings of the IEEE/CVF International Conference on Computer Vision, pp. 913-922. 2021.
\bibitem{2}
Wang, Yue, Vitor Campagnolo Guizilini, Tianyuan Zhang, Yilun Wang, Hang Zhao, and Justin Solomon. "Detr3d: 3d object detection from multi-view images via 3d-to-2d queries." In Conference on Robot Learning, pp. 180-191. PMLR, 2022.
\bibitem{3}
Reiher, Lennart, Bastian Lampe, and Lutz Eckstein. "A sim2real deep learning approach for the transformation of images from multiple vehicle-mounted cameras to a semantically segmented image in bird’s eye view." In 2020 IEEE 23rd International Conference on Intelligent Transportation Systems (ITSC), pp. 1-7. IEEE, 2020.
\
\bibitem{4}
Philion, Jonah, and Sanja Fidler. "Lift, splat, shoot: Encoding images from arbitrary camera rigs by implicitly unprojecting to 3d." In Computer Vision–ECCV 2020: 16th European Conference, Glasgow, UK, August 23–28, 2020, Proceedings, Part XIV 16, pp. 194-210. Springer International Publishing, 2020.
\bibitem{5}
Hu, Anthony, Zak Murez, Nikhil Mohan, Sofía Dudas, Jeffrey Hawke, Vijay Badrinarayanan, Roberto Cipolla, and Alex Kendall. "Fiery: Future instance prediction in bird's-eye view from surround monocular cameras." In Proceedings of the IEEE/CVF International Conference on Computer Vision, pp. 15273-15282. 2021.

\bibitem{6}
Pan, Bowen, Jiankai Sun, Ho Yin Tiga Leung, Alex Andonian, and Bolei Zhou. "Cross-view semantic segmentation for sensing surroundings." IEEE Robotics and Automation Letters 5, no. 3 (2020): 4867-4873.
\bibitem{7}
Yang, Weixiang, Qi Li, Wenxi Liu, Yuanlong Yu, Yuexin Ma, Shengfeng He, and Jia Pan. "Projecting your view attentively: Monocular road scene layout estimation via cross-view transformation." In Proceedings of the IEEE/CVF conference on computer vision and pattern recognition, pp. 15536-15545. 2021.
\bibitem{c1}
Chai, Yuning, Benjamin Sapp, Mayank Bansal, and Dragomir Anguelov. "Multipath: Multiple probabilistic anchor trajectory hypotheses for behavior prediction." arXiv preprint arXiv:1910.05449 (2019).
\bibitem{c2}
Fang, Liangji, Qinhong Jiang, Jianping Shi, and Bolei Zhou. "Tpnet: Trajectory proposal network for motion prediction." In Proceedings of the IEEE/CVF Conference on Computer Vision and Pattern Recognition, pp. 6797-6806. 2020.
\bibitem{c3}
Liang, Ming, Bin Yang, Rui Hu, Yun Chen, Renjie Liao, Song Feng, and Raquel Urtasun. "Learning lane graph representations for motion forecasting." In Computer Vision–ECCV 2020: 16th European Conference, Glasgow, UK, August 23–28, 2020, Proceedings, Part II 16, pp. 541-556. Springer International Publishing, 2020.
\bibitem{c4}
Lopez, Pablo Alvarez, Michael Behrisch, Laura Bieker-Walz, Jakob Erdmann, Yun-Pang Flötteröd, Robert Hilbrich, Leonhard Lücken, Johannes Rummel, Peter Wagner, and Evamarie Wießner. "Microscopic traffic simulation using sumo." In 2018 21st international conference on intelligent transportation systems (ITSC), pp. 2575-2582. IEEE, 2018.
\bibitem{c5}
Dosovitskiy, Alexey, German Ros, Felipe Codevilla, Antonio Lopez, and Vladlen Koltun. "CARLA: An open urban driving simulator." In Conference on robot learning, pp. 1-16. PMLR, 2017.
\bibitem{d1}
Wang, Shihao, Yingfei Liu, Tiancai Wang, Ying Li, and Xiangyu Zhang. "Exploring Object-Centric Temporal Modeling for Efficient Multi-View 3D Object Detection." arXiv preprint arXiv:2303.11926 (2023).

\bibitem{nuplan}
\textit{nuScenes: A multimodal dataset for autonomous driving},
\url{https://www.nuscenes.org/}.
\end{thebibliography}
\end{document}